\documentclass[letterpaper, 10pt, conference]{IEEEtran}
\IEEEoverridecommandlockouts
\usepackage{float}
\usepackage{filecontents}
\usepackage{cite}
\usepackage{amsmath,amssymb,amsfonts}
\usepackage{balance}
\usepackage{algorithmic}
\usepackage{algorithm}
\usepackage{graphicx}
\usepackage{textcomp}
\usepackage{enumerate}
\usepackage{subfigure}
\usepackage{relsize}
\usepackage{dblfloatfix}
\usepackage[dvipsnames]{xcolor}
\usepackage{tikz}
\usetikzlibrary{calc}
\usetikzlibrary{positioning, shapes, arrows.meta}

\DeclareMathOperator*{\argmax}{\arg\!\max}

\newcommand{\ignore}[1]{}

\newcommand{\etal}{\textit{et al.}}
\newcommand{\Tasks}{\mathsf{T}}

\newcommand{\SEthree}{\mathrm{SE}(3)}

\makeatletter
\newcommand{\superimpose}[3][\mathord]{#1{\mathpalette\superimpose@{{#2}{#3}}}}
\newcommand{\superimpose@}[2]{\superimpose@@{#1}#2}
\newcommand{\superimpose@@}[3]{%
  \ooalign{%
    \hfil$\m@th#1#2$\hfil\cr
    \hfil$\m@th#1#3$\hfil\cr
  }%
}
\makeatother

\newcommand{\bigcross}{\superimpose[\mathrel]{}{\mathlarger{{\mathlarger{\times}}}}}

\usepackage{pgfplots}
\usepackage{pgfplotstable}
\usepackage{pgfplots}
\usepgfplotslibrary{groupplots} 
\pgfplotsset{compat=1.17}
\pgfplotscreateplotcyclelist{colorblind friendly}{
    {blue, thick}, 
    {orange, thick}, 
    {cyan, thick} 
    {purple, thick}, 
    {green, thick}, 
    {red, thick}, 
    {yellow, thick}, 
}

\pgfplotscreateplotcyclelist{colorblind-friendly}{
    {blue, solid, line width=1pt},
    {orange, dashed, line width=1pt},
    {green, dashdotted, line width=1pt},
    {purple, densely dotted, line width=1pt},
    {red, densely dashed, line width=1pt},
    {brown, densely dashdotted, line width=1pt},
    {pink, loosely dashed, line width=1pt},
    {gray, loosely dashdotted, line width=1pt}
}

\def\BibTeX{{\rm B\kern-.05em{\sc i\kern-.025em b}\kern-.08em
    T\kern-.1667em\lower.7ex\hbox{E}\kern-.125emX}}
\begin{document}

\title{\LARGE \bf
A Systematic Robot Design Optimization Methodology with Application to Redundant Dual-Arm Manipulators
}

\author{Dominic Guri$^{1}$ and George Kantor$^{1}$
\thanks{This work was supported in part by NSF Robust Intelligence 1956163, NSF/USDA-NIFA AIIRA AI Research Institute 2021-67021-35329, and USDA-NIFA LEAP 2024-51181-43291.}
\thanks{$^{1}$Dominic Guri and George Kantor are with the Robotics Institute, Carnegie Mellon University, Pittsburgh 15206 USA 
        {\tt\small \{dguri, kantor\}@cs.cmu.edu}}%
}

\maketitle
\thispagestyle{empty}
\pagestyle{empty}

\begin{abstract}
One major recurring challenge in deploying manipulation robots is determining the optimal placement of manipulators to maximize performance. This challenge is exacerbated in complex, cluttered agricultural environments of high-value crops, such as flowers, fruits, and vegetables, that could greatly benefit from robotic systems tailored to their specific requirements. However, the design of such systems remains a challenging, intuition-driven process, limiting the affordability and adoption of robotics-based automation by domain experts like farmers. To address this challenge, we propose a four-part design optimization methodology for automating the development of task-specific robotic systems. This framework includes (a) a robot design model, (b) task and environment representations for simulation, (c) task-specific performance metrics, and (d) optimization algorithms for refining configurations. We demonstrate our framework by optimizing a dual-arm robotic system for pepper harvesting using two off-the-shelf redundant manipulators. To enhance performance, we introduce novel task metrics that leverage self-motion manifolds to characterize manipulator redundancy comprehensively. Our results show that our framework achieves simultaneous improvements in reachability success rates and improvements in dexterity.
Specifically, our approach improves reachability success by at least 14\% over baseline methods and achieves over 30\% improvement in dexterity based on our task-specific metric.
\end{abstract}



\section{INTRODUCTION}
Optimal robot placement is a critical yet often overlooked aspect of robot automation design and optimization. It can mean the difference between successfully executing a task and requiring costly custom solutions. In agriculture --- particularly in high-value specialty crops such as flowers, vegetables, and fruits --- better robot placement can lower barriers to adoption, making versatile robotic manipulation systems more accessible to practitioners like farmers and researchers.

In this work, we develop a systematic approach to optimal robot design, demonstrating its effectiveness by examining how reconfiguring robots can enhance task-specific performance. Our focus is improving redundant manipulators' versatility in cluttered environments, particularly in agricultural settings. Robot configuration optimization is inherently complex, time-consuming, and often counterintuitive, even for experts. These challenges limit performance and slow the broader adoption of robotic automation. By introducing a structured optimization methodology for modular robot design, we aim to overcome these barriers, enabling more efficient and adaptable robotic solutions.

\section{RELATED WORKS}
Robot design optimization is a bilevel optimization problem consisting of an inner problem that determines the optimal performance for a given design candidate and an outer problem that iteratively refines the design to maximize overall performance. This approach has been widely applied in the design of both manipulators~\cite{chenEnumeratingNonisomorphicAssembly1993,chenDeterminingTaskOptimal1995,chungTaskBasedDesign1997,chocronEvolutionaryAlgorithmsKinematic1997,sakkaOptimalDesignConfigurations2001,mosadeghzadDynamicModelingStability2012,icerTaskdrivenAlgorithmConfiguration2016} and mobile robots~\cite{wooOptimalDesignNew2007,chocronEvolvingModularRobots2007,schenkerReconfigurableRobotsAllterrain2000,udengaardDesignOmnidirectionalMobile2008,liParameterDesignOptimization2009,kimOptimalDesignKinetic2012}. Formulating a design optimization problem requires answering several key questions: What constitutes the robot model? How is performance measured? What physical and operational constraints must be considered? 

Because assessing robot design performance often requires running a simulation that can be a simple reachability test or a complete dynamic simulation~\cite{schenkerReconfigurableRobotsAllterrain2000,icerTaskdrivenAlgorithmConfiguration2016,huModularRobotDesign2022}, evaluating the inner optimization problem can be computationally expensive, and non-differentiable. Constraints such as reachability, joint angle limits, torque limits, and obstacle avoidance further restrict the design search space. As a result, gradient-free and meta-heuristic search algorithms are commonly employed in the literature. Early work in robot design, for instance, made extensive use of customized genetic algorithms \cite{parkerInverseKinematicsRedundant1989, chenEnumeratingNonisomorphicAssembly1993, chenDeterminingTaskOptimal1995, chocronEvolutionaryAlgorithmsKinematic1997}, an approach that remains popular in recent studies \cite{mosadeghzadDynamicModelingStability2012, kulzOptimizingModularRobot2023}. Other optimization techniques, including Simulated Annealing \cite{paredisKinematicDesignSerial1993} and bio-inspired algorithms such as Harris Hawks Optimization (HHO), Grey Wolf Optimization (GWO), and Particle Swarm Optimization (PSO), have also proven effective in navigating the complex search spaces typical of robot design~\cite{galan-uribeKinematicOptimization6DOF2022}.

\begin{figure} 
    \centering
    \begin{tikzpicture}[      
            every node/.style={anchor=south west,inner sep=0pt},
            x=1mm, y=1mm,
          ]   
         \node (fig1) at (0,0)
           {\includegraphics[scale=0.6]{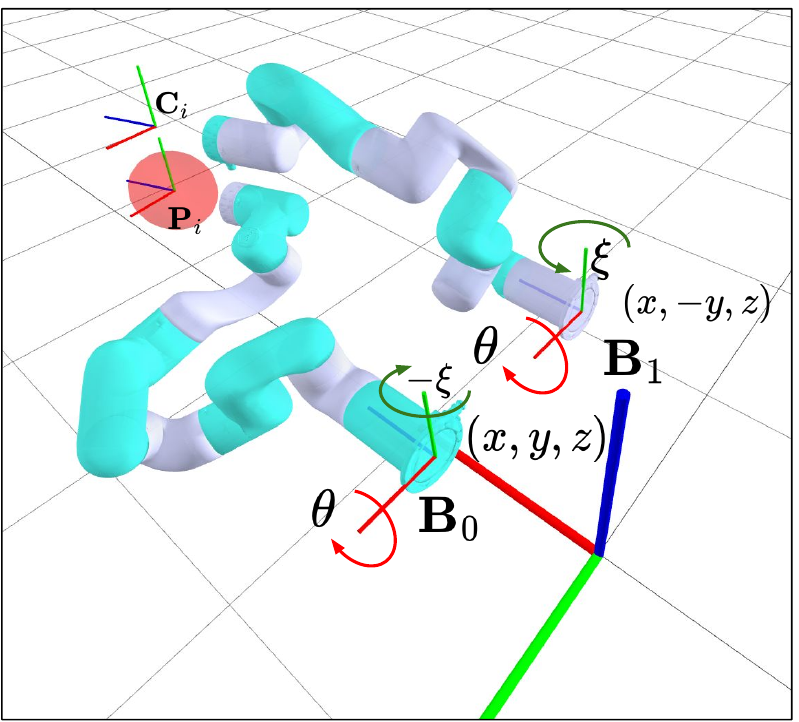}};
         \node[draw, line width=0.5mm, color=black] (fig2) at (3,3)
           {\includegraphics[scale=0.2]{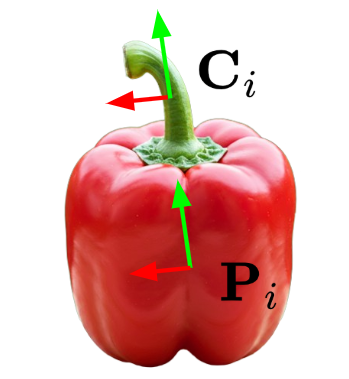}};  
    \end{tikzpicture}
    \caption{The figure above illustrates the human-like dual-arm robot design, parameterized by $\rho = (x, y, z, \theta, \xi)$. The end-effectors are assigned to reach frames $\mathbf{P}_i$ and $\mathbf{C}_i$, which correspond to the pepper center and peduncle for grasping and cutting, respectively.}\label{fig:robot_model-description}
\end{figure}

The challenge of determining the optimal robot placement is crucial in both industrial~\cite{pamanes-garciaCriterionOptimalPlacement1990,zeghloulMulticriteriaOptimalPlacement1993,abdel-malekPlacementRobotManipulators2004} and surgical~\cite{shiWorkingPoseLayout2020} applications. A closely related problem is workpiece placement optimization, which is also widely studied in these domains~\cite{balciOptimalWorkpiecePlacement2023,zelechowskiPatientPositioningVisualising2022,zelechowskiAutomaticPatientPositioning2023}. Research in this area generally follows two main approaches: (1) defining a design objective based on a dexterity measure evaluated at task-relevant end-effector poses~\cite{abdel-malekPlacementRobotManipulators2004,balciOptimalWorkpiecePlacement2023}, or (2) constructing a comprehensive capability map of the robot, ranging from simple reachable volume estimation to augmented models incorporating dexterity metrics~\cite{zachariasCapturingRobotWorkspace2007, zachariasCAPABILITYMAPTOOL2014, zhangConfigurationLayoutPose2021,sandakalumInvReachNetDeciding2022}. Some of the earliest placement optimization studies prioritized avoiding joint limits and obstacle collisions using heuristic methods~\cite{pamanes-garciaCriterionOptimalPlacement1990,zeghloulMulticriteriaOptimalPlacement1993}. Abdel-Malek and Yu~\cite{abdel-malekPlacementRobotManipulators2004} introduced a novel dexterity measure that combined manipulability~\cite{yoshikawaManipulabilityRoboticMechanisms1985} with joint limit constraints, ensuring target poses remained within a threshold distance from singularity surfaces. More recent works have explored various optimization techniques: Tugal \etal~\cite{tugalManipulationOptimumLocations2022} employed heuristic search to optimize robot placement for maximum force transmissibility, and Balci \etal~\cite{balciOptimalWorkpiecePlacement2023} applied nonlinear optimization to maximize manipulability and minimize torque in a surface grinding task.

Workspace characterization has also evolved significantly. Early works, such as Seraji~\cite{serajiReachabilityAnalysisBase1995}, focused on fitting the workpiece within the robot's reachable volume. Over time, researchers incorporated dexterity considerations into workspace modeling. Zacharias \etal~\cite{zachariasCapturingRobotWorkspace2007} introduced \textit{reachability spheres} to capture the directionality of feasible manipulation actions. Vahrenkamp \etal~\cite{vahrenkampRobotPlacementBased2013} developed an inverted distribution representation to model robot base pose distributions for reaching specific workspace objects. In a later study~\cite{vahrenkampRepresentingRobotsWorkspace2015}, they further refined this approach by incorporating an augmented Jacobian representation that accounts for directionality, obstacles, and manipulator kinematics. Zhang \etal~\cite{zhangConfigurationLayoutPose2021} extended this concept by considering multiple inverse kinematics (IK) solutions per target pose to capture the full range of feasible joint configurations.

\begin{figure}
    \centering
    \begin{tikzpicture}[
    font=\sffamily,
    node distance=0.5cm and 1.8cm,
    block/.style = {rectangle, draw, fill=white, text centered, minimum height=1.75em, minimum width=35mm},
    arrow/.style={-{Stealth[length=3mm,width=2mm]}, thick, draw=black}
    ]
        \node (robot) [block] {Robot Design};
        \node (task) [block, right=of robot] {Task/Environment};
        \node (metric) [block, below=of $(robot)!0.5!(task)$] {Task Metric};
        \node (opt) [block, below=of metric] {Optimization};
        \node (design) [block, draw=none, below=of opt, text width=30mm, font=\bfseries\itshape] {Optimal Design};
        \draw [arrow] (robot) |- (metric);
        \draw [arrow] (task) |- (metric);
        \draw [arrow] (metric) -- (opt);
        \draw [arrow] (opt) -- (design);
    \end{tikzpicture}
    \caption{
    The robot design optimization framework requires four element:
    (1) a robot model parameterized by optimization decision variables,
    (2) task and environment representation to model the physical constraints and the expected robot operations,
    (3) a task metric -- a function that evaluates the performance of a given robot design in a simulated task , and
    (4) an optimizer for updating robot parameters.}
    \label{fig:design-pipeline}
\end{figure}
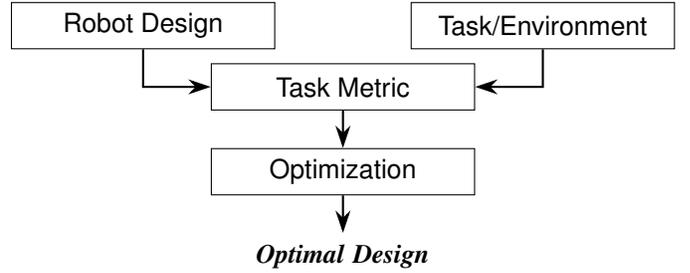

\subsection{Redundant Manipulators}
In this work, we focus on redundant manipulators --- robot arms with more degrees of freedom than strictly required for a given task --- because they provide an infinite number of joint configurations for reaching a target; this is particularly advantageous in cluttered environments. However, the abundance of feasible joint configurations makes evaluating dexterity-based task metrics more challenging. Typically, secondary objectives such as collision avoidance and dexterity maximization are incorporated as auxiliary tasks within numerical inverse kinematics (IK) solvers~\cite{dasInverseKinematicAlgorithms1988,escandeHierarchicalQuadraticProgramming2013,havilandManipulatorDifferentialKinematics2023}. While these solvers are versatile and relatively straightforward to implement, they do not guarantee convergence --- a limitation that is further worsened in cluttered conditions. Consequently, techniques such as solver restarts or employing multiple solvers in parallel are often recommended to improve reliability~\cite{havilandManipulatorDifferentialKinematics2023}. 

To overcome the limitations of numerical inverse kinematics in assessing robot performance, we leverage self-motion manifolds (SMMs). Originally introduced by Burdick\cite{burdickInverseKinematicsRedundant1989}, a self-motion manifold, $\mathcal{S}_\mathbf{x}$, represents the set of all joint configurations that achieve the same end-effector pose $\mathbf{x}$, formally defined as:
\begin{align}
    \mathcal{S}_\mathbf{x} =
    \big\{ 
    ~\mathbf{q} \in \mathcal{C} ~|~ \mathbf{x} = \phi(\mathbf{q})~
    \big\} \label{eqn:smm}
\end{align}
Rather than computing dexterity measures through numerical inverse kinematics, we evaluate them directly on the self-motion manifolds --- this ensures that the best possible performance is identified whenever it exists.

\begin{figure}
    \centering
    \includegraphics[width=1\linewidth]{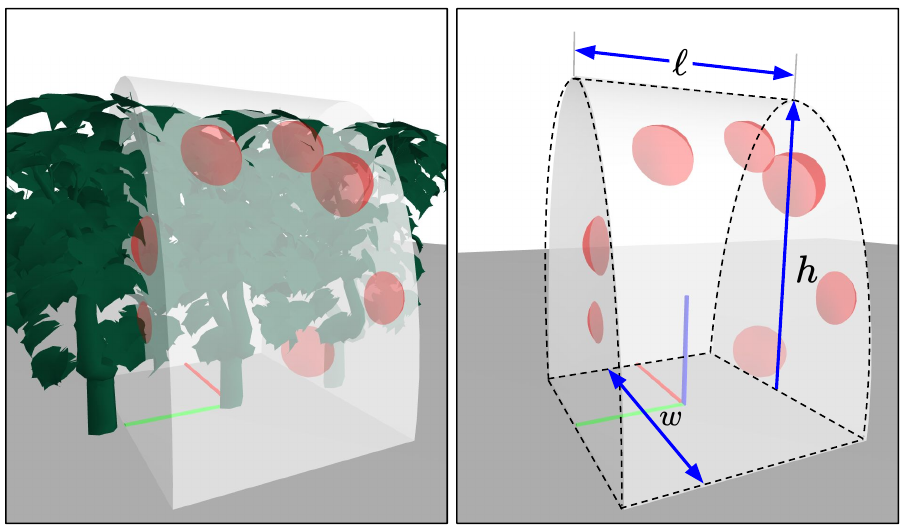}
    \caption{
    The diagram above illustrates the ridge canopy environment model, with pepper models sampled within the canopy volume. For the experiments, $h=0.8$m and $w=\ell=0.6$m, and the canopy is located at $(1,0,0)$, extending along the $y$-axis.}\label{fig:env-model}
\end{figure}

Almarkhi and Maciejewski \cite{almarkhiMaximizingSizeSelfMotion2019} computed SMMs by tracking the change of the end-effector pose when only one joint was perturbed. They also proposed an algorithm for identifying the largest SMMs by tracking and classifying singularities based on rank, noting that the rank of the singularity increases when multiple SMMs intersect. According to Groom \etal \cite{groomRealtimeFailuretolerantControl1999}, configurations from the largest usable section of an SMM enhance fault tolerance by providing the largest maneuverable joint configuration space without compromising the end-effector. Other methods for computing SMMs include a grid search approach by Peidr\'o \etal~\cite{peidroMethodBasedVanishing2018}, which identifies SMM configurations by discretizing the joint space. Wu \etal~\cite{wuNovelMethodComputing2023} adopted a hypercube grid representation of the joint configuration space, combined with cellular automata rules to compute SMMs.

\subsection{Contributions}
In this work, we introduce:
\begin{enumerate}[(a)]
    \item A systematic framework for formulating robot design optimization problems. Our framework consists of (\textit{i}) a robot model, (\textit{ii}) task and environment representation, (\textit{iii}) a task metric, and (\textit{iv}) an optimizer.
    \item A new class of dexterity measures based on self-motion manifolds designed to be exhaustive in assessing performance.
    \item A demonstration of the framework's effectiveness by solving for the optimal mounting configurations of a dual-arm system optimized for pepper harvesting. 
\end{enumerate}
Through these contributions, we aim to advance the development of automated robot design and reconfiguration tools that reduce the specialized knowledge required to customize robots for better performance.

The remainder of this manuscript is organized as follows: Section~\ref{sec:ii:methodology} details the design methodology within the proposed four-element framework. Section~\ref{sec:iii:example} introduces the pepper harvesting design problem, while Section~\ref{sec:iv:experiments} describes the experimental setup for benchmarking the optimized and baseline solutions. The results are presented in Section~\ref{sec:v:results}, followed by the conclusion in Section~\ref{sec::conclusion}.

\section{CONFIGURATION OPTIMIZATION METHODOLOGY FOR REDUNDANT MANIPULATORS}\label{sec:ii:methodology}

We present a systematic framework for formulating the robot design optimization problem, specifically tailored to the challenges of cluttered environments. Central to our approach is a novel task metric formulation based on self-motion manifolds (SMMs), which comprehensively evaluate the best possible performance of a redundant manipulator. This methodology ensures consistent performance evaluation across designs, ultimately leading to improved design optimization and enhanced robotic capabilities for complex tasks.

\subsection{Design Methodology}
Our design methodology is built on iteratively refining the robot design based on its performance in executing tasks within a simulated environment. The design problem is structured around four key elements, as illustrated in Figure~\ref{fig:design-pipeline}: (a) the robot model, (b) the task and environment representation, (c) the task metric, and (d) the optimizer. The success of the design process, assuming a high-quality optimizer, hinges on several factors: the accuracy of the robot model, the fidelity of the task and environment representation in capturing real-world conditions, the effectiveness of the control policy employed during simulation, and the robustness of the task metric in evaluating robot performance. However, as the fidelity of the robot, task, and environment models increases to reflect real-world conditions better, the computational cost of simulation and, consequently, the optimization process also increases significantly.

\subsubsection{Robot Model}
A robot model typically includes the robot's kinematic, geometric (visualization and collision), and dynamic properties. For our purposes, the model is defined as the subset of these properties necessary to simulate the task's execution. Additionally, the robot model must be parameterized by the design variables, which are kinematic, geometric, or dynamic properties that affect the robot's performance. These design variables serve as the decision variables in the robot optimization problem. 

Our robot model is defined by its forward kinematic function and collision geometry to optimize the base pose of a manipulator. The forward kinematic function is defined as
$\phi:\mathcal{C}\times\Omega\to\mathcal{X}$, where $\mathcal{C}$ and $\mathcal{X}$ are the joint configuration space and the task space, respectively, and $\Omega$ is the design variable search space -- a set of allowable robot bases in $\SEthree$, generally defined by constraints on position and orientation of the base. Hence, for a given design $\rho\in\Omega\subseteq\SEthree$ and joint configuration $\mathbf{q}\in\mathcal{C}$, the end-effector pose $\mathbf{x}\in\mathcal{X}$ is determined as: $\mathbf{x}=\phi(\mathbf{q};\rho)$. 

\subsubsection{Task \& Environment Representation} 
The task and environment representations are defined together since the environment imposes constraints on the execution of the task. 

\paragraph{The environment} At a minimum, the environment is represented by a set $\mathcal{O} = \{\mathbf{o}_0, \mathbf{o}_1, \dots\}$, where each element corresponds to the geometry of an obstacle in the environment. Higher-fidelity representations may also include semantic details about the obstacles, essential for advanced control policies that can effectively exploit environmental affordances.

\paragraph{Tasks} Manipulation tasks range from simple reaching tasks to complex trajectory tracking and dynamic motions, such as those required in industrial applications. Mayer \etal~\cite{mayerCoBRAComposableBenchmark2022} provide a comprehensive classification of manipulation tasks, particularly for industrial contexts. We will assume a simple reaching task where the robot's end-effector is required to reach a set of specific end-effector poses, $\mathbf{x}_i$, within a defined tolerance, $\delta_i$.

We adopt a task-environment representation convention inspired by Mayer \etal~\cite{mayerCoBRAComposableBenchmark2022}, who define a task as three-tuple, $\mathsf{T} = \langle \mathbf{x}_i, \delta_i, \mathcal{O} \rangle$, consisting of the target pose, tolerance, and obstacle set. For simplicity, we drop the tolerance and represent the task as a two-tuple, $\mathsf{T} = \langle \mathbf{x}_i, \mathcal{O} \rangle$, with the default tolerance value defined in the inverse kinematics solver.

\subsubsection{Task Metric}
The task metric is a function $\tau(\rho; \mathsf{T})$ that assigns a real-valued numerical score—a measure of performance—to a robot design $\rho$ based on its execution of a simulated task $\mathsf{T}$. Task metrics can be defined in various ways, ranging from simple binary reachability tests to more complex measures such as trajectory tracking success probabilities.

In this work, we focus on dexterity measures like manipulability. The objective is to optimize the placement of the manipulator so that the target poses are highly manipulable, i.e., the capacity for dexterous manipulation is as high as possible. We explore the use of redundant manipulators with a task redundancy of 1, operating in cluttered environments where determining reachability poses unique challenges. As the level of clutter increases, conventional IK methods fail, yet redundancy is supposed to increase the likelihood of obtaining a solution if one exists. 
To address IK challenges for redundant manipulators, we compute the SMM $\mathcal{S}_\mathbf{x}$ (Equation~\ref{eqn:smm}) for a given end-effector pose $\mathbf{x}$, then define the task metric for reaching $\mathbf{x}$ as $\tau(\rho; \mathbf{x})$, as a three-stage function:

\begin{align}
    \tau(\rho;\mathsf{T}) 
        &= \mathsf{Generate}~\mathcal{S}_\mathbf{x} 
        \to \mathsf{Filter}
        \to \mathsf{Criterion} \label{eqn:3stage-task-metric}
\end{align}
where:
\begin{itemize} 
    \item $\mathsf{Generate}~\mathcal{S}_\mathbf{x}$: Computes the self-motion manifold (SMM), capturing all valid joint configurations $\mathbf{q} \in \mathcal{C}$ that achieve the specified end-effector pose $\mathbf{x}$. 
    \item $\mathsf{Filter}$: Removes joint configurations from $\mathcal{S}_\mathbf{x}$ that violate constraints such as joint limits, self-collisions, and obstacle collisions. The resulting filtered set of configurations is denoted as $\mathcal{U}_\mathbf{x}$. 
    \item $\mathsf{Criterion}$: Evaluates the performance based on the filtered set $\mathcal{U}_\mathbf{x}$. For example, the criterion function could compute the configuration that maximizes manipulability or other dexterity measures.
\end{itemize}
This approach enhances consistency in comparing robot designs by ensuring the task metric evaluates all feasible joint configurations while incorporating task-specific performance criteria.

We restrict our analysis to task redundancies of dimensionality 1, such as a 7DOF manipulator performing a 6DOF task, where we have methods of exhaustively computing the SMM. We define the maximum manipulability criterion $c(\mathcal{U}_\mathbf{x})$ which evaluates the Yoshikawa manipulability measure $\mu(\mathbf{q})$ (Equation~\ref{eqn:yoshikawa}) for each configuration in $\mathcal{U}_\mathbf{x}$ and returns the highest score.

To provide a more comprehensive manipulability task metric, we introduce the concept of \textit{manipulability density}:
\begin{align}
    \tilde{\mu}(\mathcal{U}_\mathbf{x}) &= \frac{1}{\vert \mathcal{U}_\mathbf{x} \vert}\int_{\mathcal{U}_\mathbf{x}}~\mu(\mathbf{q})~d\mathbf{q} \label{eqn:manip-density} \\
    \mu(\mathbf{q}) &= \sqrt{\det(\mathbf{J(q)}\mathbf{J^\top(q)})} \label{eqn:yoshikawa}
\end{align}
which characterizes manipulability across the entire set $\mathcal{U}_\mathbf{x}$. This approach captures the overall manipulability potential, addressing scenarios where the configuration with the highest manipulability may not be feasible due to varying clutter conditions.

\subsubsection{Optimization}
The underlying design optimization problem formulation depends on the nature of the operating conditions.
If the environment and task are known and fixed, the optimal design $\rho^*$ maximizes the task metric as follows 
\begin{align} 
    \rho^* &= \argmax_{\rho \in \Omega}~\tau(\rho; \mathsf{T}). 
    \label{eqn:design-task-optimization}
\end{align}
However, if the task cannot be easily modeled and only samples form a task distribution can be provided, the design optimization challenge is an expected performance maximization problem:
\begin{align} 
    \rho^* &= 
    \argmax_{\rho \in \Omega}
    {}~\int_\mathcal{T} 
        \tau(\rho; \mathsf{T}) P(\varsigma)~d\varsigma
        \label{eqn:design-expectation-maximization}
\end{align}
where $P(\varsigma)$ is the probability of sampling a task $\varsigma$ from the task distribution.

In our approach, the task metric $\tau(\rho; \mathsf{T})$ is non-differentiable and computationally expensive. Consequently, we employ gradient-free optimization methods. Specifically, we use Particle Swarm Optimization (PSO) due to its demonstrated stability and performance guarantees in handling non-convex, high-dimensional search spaces~\cite{kennedyParticleSwarmOptimization1995, xuConvergenceAnalysisParticle2018, gadParticleSwarmOptimization2022}.

We discretize the search space to ensure the final design is manufacturable, restricting it to positions and angles compatible with our mobile platform. Discretization not only ensures that physical construction is feasible but also enables memoization~\cite{ahoDesignAnalysisComputer1974}, i.e., improving computational efficiency by caching objective function results for previously queried design parameters to avoid redundant computations.

\section{PEPPER HARVESTING EXAMPLE}\label{sec:iii:example}
In this section, we demonstrate the utility of our design methodology using Hefty~\cite{guriHeftyModularReconfigurable2024} (shown in Figure~\ref{fig:real-conditions-model}), a modular and reconfigurable robot designed to be a utility platform for a wide range of mobile manipulation research in agriculture and field robotics. The specific problem we address is determining the optimal placement of two 7DOF xArm7 manipulators~\cite{UFACTORYXArmUFACTORY} to maximize their shared reach within the plant canopy for pepper harvesting. One manipulator is tasked with grasping the pepper, while the other is responsible for cutting the peduncle. The four elements of the design methodology are defined as follows:

\begin{figure}
    \centering
    \includegraphics[width=\linewidth]{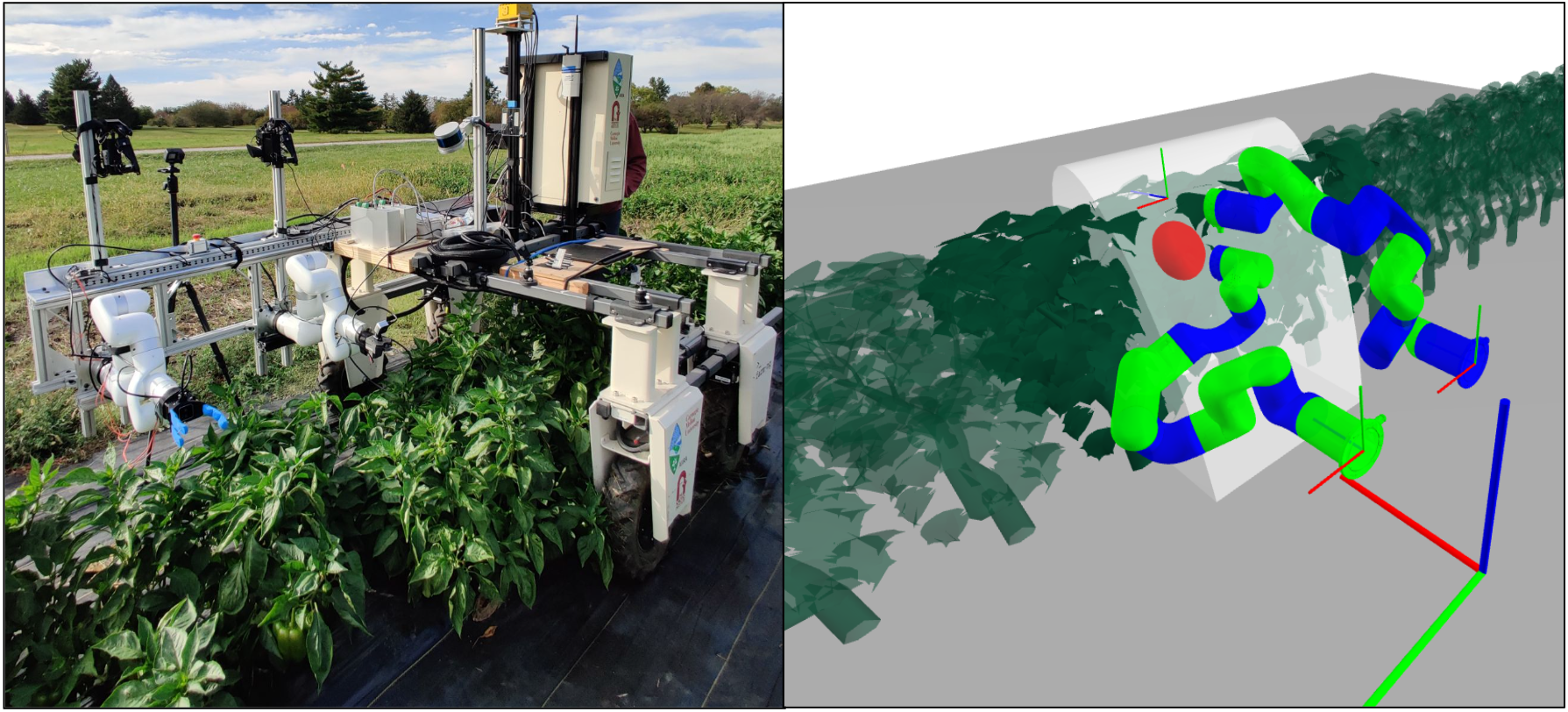}
    \caption{The first image shows the reference \textit{expert} design that was previously deployed for a teleoperated system for human demonstration data collection. The second image illustrates the ``ridge'' canopy model that represents pepper plot section assgined to the harvesting robot.}\label{fig:real-conditions-model}
\end{figure}

\subsubsection{Robot Model}
We define a robot model $\Phi: \mathcal{C} \times \Omega \to \mathcal{X}$, which maps the joint space of both manipulators and that of the design variable to the combined workspace of the manipulators. Specifically, the robot model computes the end-effector poses, given the joint configurations and the design variable $\rho \in \Omega$. The design variable $\rho$ is used to compute the base transformations $\mathbf{B}_0(\rho)$ and $\mathbf{B}_1(\rho)$ in $\SEthree$ as follows:
\begin{align}
    \begin{pmatrix}
        \mathbf{x}_0 \\ \mathbf{x}_1
    \end{pmatrix} = 
    \Phi(\mathbf{q}_0, \mathbf{q}_1; \rho) = 
    \begin{pmatrix}
            \mathbf{B}_1(\rho) \cdot \phi(\mathbf{q}_0) \\ 
            \mathbf{B}_0(\rho) \cdot \phi(\mathbf{q}_1)
    \end{pmatrix}
    \label{eqn:dual-fk}
\end{align}
where $\phi$ represents the forward kinematics function which is assumed to be the same for both manipulators.

To simplify the design, we adopt a human-like morphology (Figure~\ref{fig:robot_model-description}) by parameterizing the base transformations as:
\begin{align}
    \mathbf{B}_0(\rho) &= 
    \mathbf{T}(x, y, z) 
        \cdot \mathbf{R}_y(\theta)
        \cdot \mathbf{R}_x(\tfrac{\pi}{2})
        \cdot \mathbf{R}_y(\tfrac{\pi}{2} - \xi) \\
    \mathbf{B}_1(\rho) &= 
    \mathbf{T}(x, -y, z) 
        \cdot \mathbf{R}_y(\theta)
        \cdot \mathbf{R}_x(\tfrac{\pi}{2})
        \cdot \mathbf{R}_y(\tfrac{\pi}{2} + \xi).
\end{align}
Here, $\rho = (x, y, z, \theta, \xi)$ defines the design variable space. $\mathbf{T}(x, y, z)$ represents a translation transformation, while $\mathbf{R}_x(\cdot)$ and $\mathbf{R}_y(\cdot)$ represent rotations about the $x$ and $y$ axes, respectively. The design maintains a shoulder-width distance of $2y$, a uniform pitch angle $\theta$ (rotation about the $y$-axis), and a mirrored camber angle $\xi$ (rotation about the $z$-axis).

\subsubsection{Task \& Environment}
\paragraph{Environment}
Figure~\ref{fig:real-conditions-model} shows a pepper farm plot where the plant canopies merge into a continuous ``ridge'' spanning the length of the plot, with all peppers nested within this ridge. To represent this environment, we model the ridge as a semi-elliptic cylinder: the major axis corresponds to the plant height, the minor axis corresponds to the plant width, and the cylinder's length spans the plot~\ref{fig:env-model}. Each pepper $i$ is assumed to lie within the ridge and is represented by a pair of frames, $\mathbf{P}_i$ and $\mathbf{C}_i$, in $\SEthree$ (Figure~\ref{fig:robot_model-description}), corresponding to the center of the pepper and the peduncle cutting point, respectively.

\paragraph{Task}
Using the environment model above, we define a reaching task as $\Tasks = \langle (\mathbf{P}_i, \mathbf{C}_i), \mathcal{O} \rangle$, where the manipulator at $\mathbf{B}_0(\rho)$ is assigned to $\mathbf{P}_i$ for grasping, and the manipulator at $\mathbf{B}_1(\rho)$ reaches $\mathbf{C}_i$ to perform the cutting action. The obstacle set $\mathcal{O}$ includes the ground plane and the ridge. The task requires the robot's geometry to avoid collisions with the ground plane, while the robot's shoulders (the immobile parts of the manipulators) must remain outside the ridge geometry.

\subsubsection{Pepper Harvesting Task Metric}
For a dual-arm system, where the arms are assigned to reach target poses $\mathbf{x}$ and $\mathbf{y}$ in $\SEthree$, we extend the task metric definition from Equation~\ref{eqn:3stage-task-metric} by returning the minimum of the respective manipulability density scores, $\tilde{\mu}(\mathcal{U}_\mathbf{x})$ and $\tilde{\mu}(\mathcal{U}_\mathbf{y})$, as shown in Equation~\ref{eqn:task-metric}. Therefore, for a task $\mathsf{T}$ consisting of a list of target pairs, the task metric is defined as:
\begin{align}
    \tau(\rho; \Tasks)
    &= \sum_{\mathbf{x},\mathbf{y}\in\Tasks}
    \min \{ \tilde{\mu}(\mathcal{U}_\mathbf{x}),~\tilde{\mu}(\mathcal{U}_\mathbf{y}) \}
    \label{eqn:task-metric}
\end{align}
where the design variable $\rho$ is used to generate the joint configuration sets $\mathcal{U}_\mathbf{x}$ and $\mathcal{U}_\mathbf{y}$, as per Equation~\ref{eqn:3stage-task-metric}. During the filtering stage, configurations that violate joint limits, cause manipulator self-collisions or collide with the ground are eliminated. If $\rho$ results in base poses that cause the shoulders to collide with each other, the ground, or the ridge, the sets $\mathcal{U}_\mathbf{x}$ and $\mathcal{U}_\mathbf{y}$ will be empty, resulting in the task metric evaluating to zero.

\subsubsection{Optimization}
We solve the design optimization problem expressed in Equation~\ref{eqn:design-task-optimization}, where we seek maximize the task metric on a fixed set of targets. We use particle swarm optimization (PSO) to search the design variable search space, as described below.
\begin{table}[ht]
    \centering
    \begin{tabular}{|c|c|c|c|}
        \hline
        \textbf{Parameter} & \textbf{Minimum} & \textbf{Step} & \textbf{Maximum} \\ \hline
        x (cm) & -5 & 2.5 & 70 \\ \hline
        y (cm) & 20 & 2.5 & 50 \\ \hline
        z (cm) & 20 & 2.5  & 100 \\ \hline
        roll (rad) & $-90^\circ$ & $5^\circ$ & $90^\circ$ \\ \hline
        pitch (rad) & $-90^\circ$ & $5^\circ$ & $90^\circ$ \\ \hline
    \end{tabular}
    \caption{Design search space discretization for constructability and memoization.}\label{tab:search-spaceparameters}
\end{table}

\begin{figure*}
    \centering
    \subfigure[Reachability success rate plot]{\includegraphics[width=0.241\textwidth]{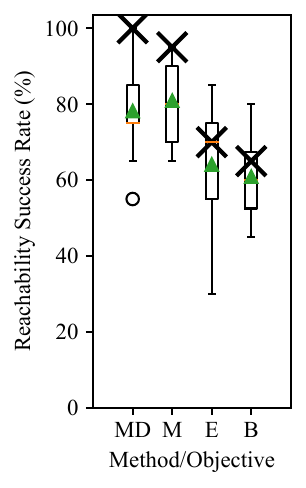}\label{fig:sr-results}}
    \subfigure[Manipulability density plot]{\includegraphics[width=0.245\textwidth]{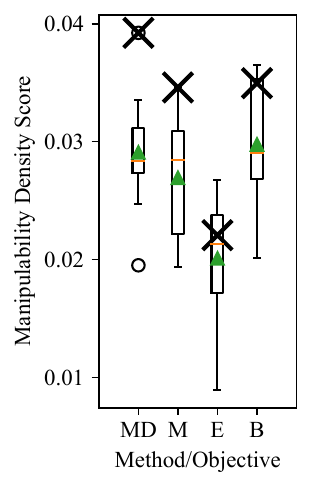}\label{fig:md-results}}
    \subfigure[Manipulability density vs. success rate plot]{\includegraphics[width=0.40\textwidth]{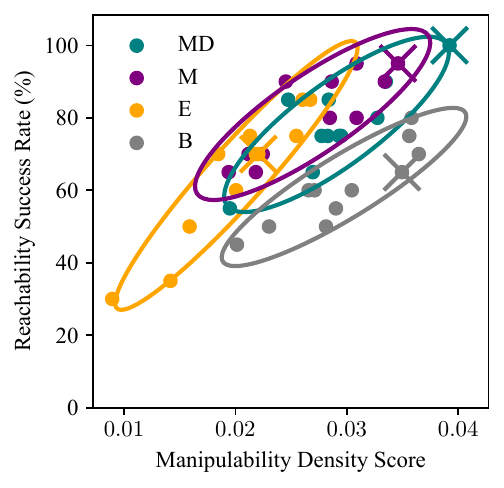}\label{fig:md-sr-results}}
    
    \caption{The plots above illustrate the design optimization performance for the manipulability (M) and manipulability density (MD) task metrics, compared to the human expert (E) and the baseline (B), which is based on a manipulability-maximizing reactive controller~\cite{havilandPurelyReactiveManipulabilityMaximisingMotion2020}. The boxplots depict the distributions of reachability success rates and manipulability density scores, evaluated on the primary optimization task dataset (marked by $\bigcross$) and 10 additional randomly sampled sets. Figure~\ref{fig:md-sr-results} presents the performance results with a 2$\sigma$-ellipse. The design methodology solutions, M and MD, consistently outperform the alternatives, whereas the expert (E) solution exhibits high variance and low dexterity scores. The baseline (B) achieves high dexterity but suffers from low reachability success. All solutions are illustrated in Figure~\ref{fig:design-results-pictures}.}\label{fig:design-results-plots}
\end{figure*}

\begin{figure*}[!b]
    \centering
    \subfigure[Manipulability Density (MD)]{\includegraphics[width=0.245\textwidth]{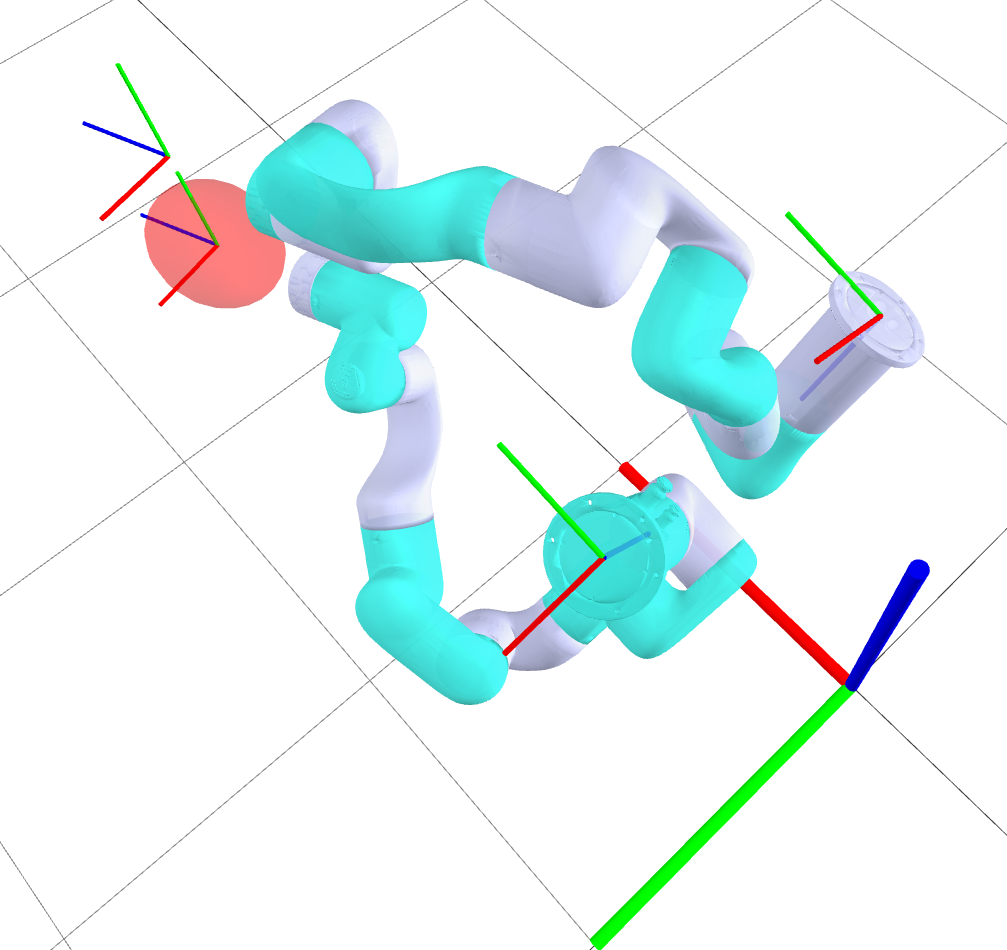}\label{fig:md-design}}
    \subfigure[Manipulability (M)]{\includegraphics[width=0.245\textwidth]{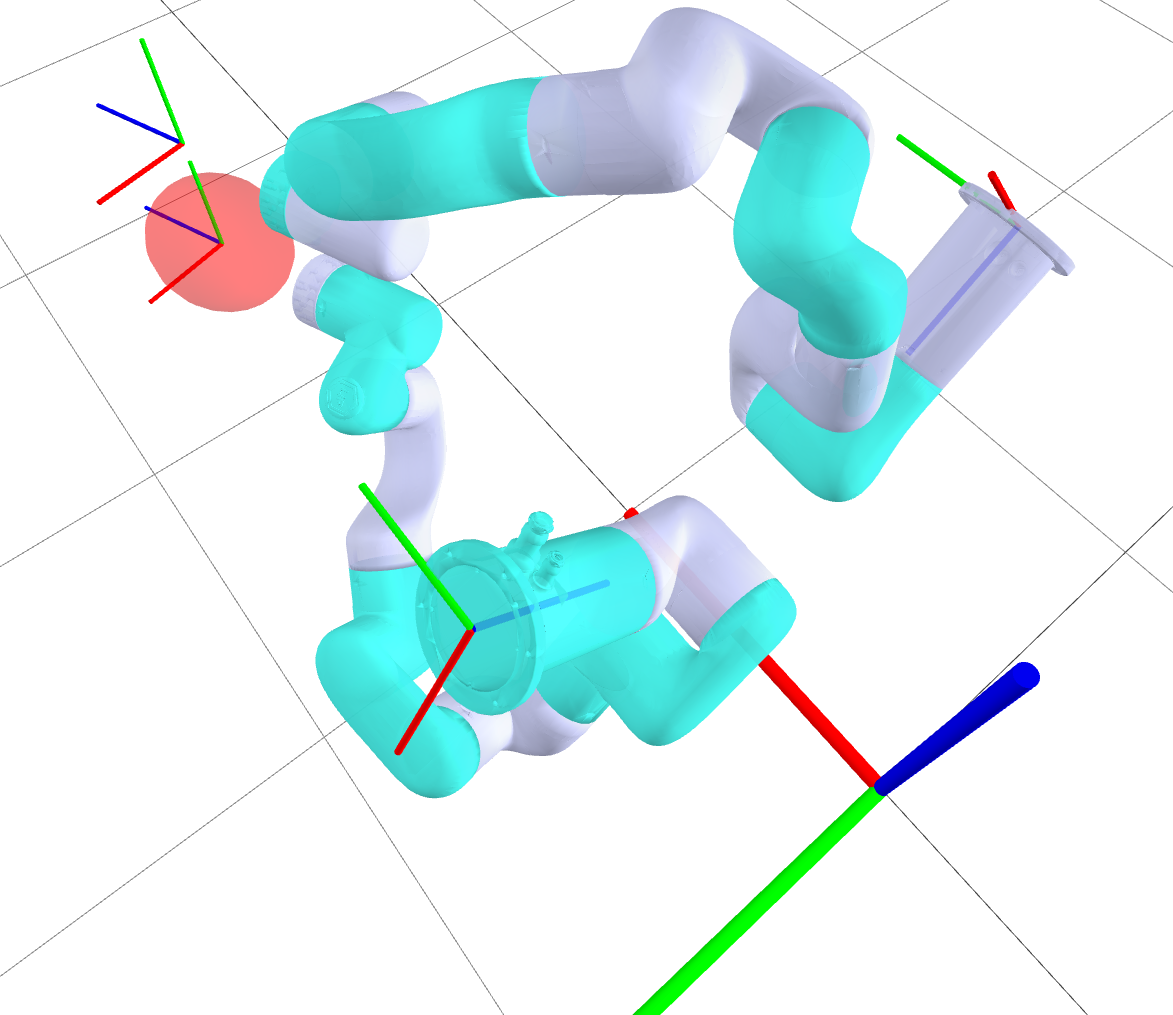}\label{fig:m-design}}
    \subfigure[Expert (E)]{\includegraphics[width=0.245\textwidth]{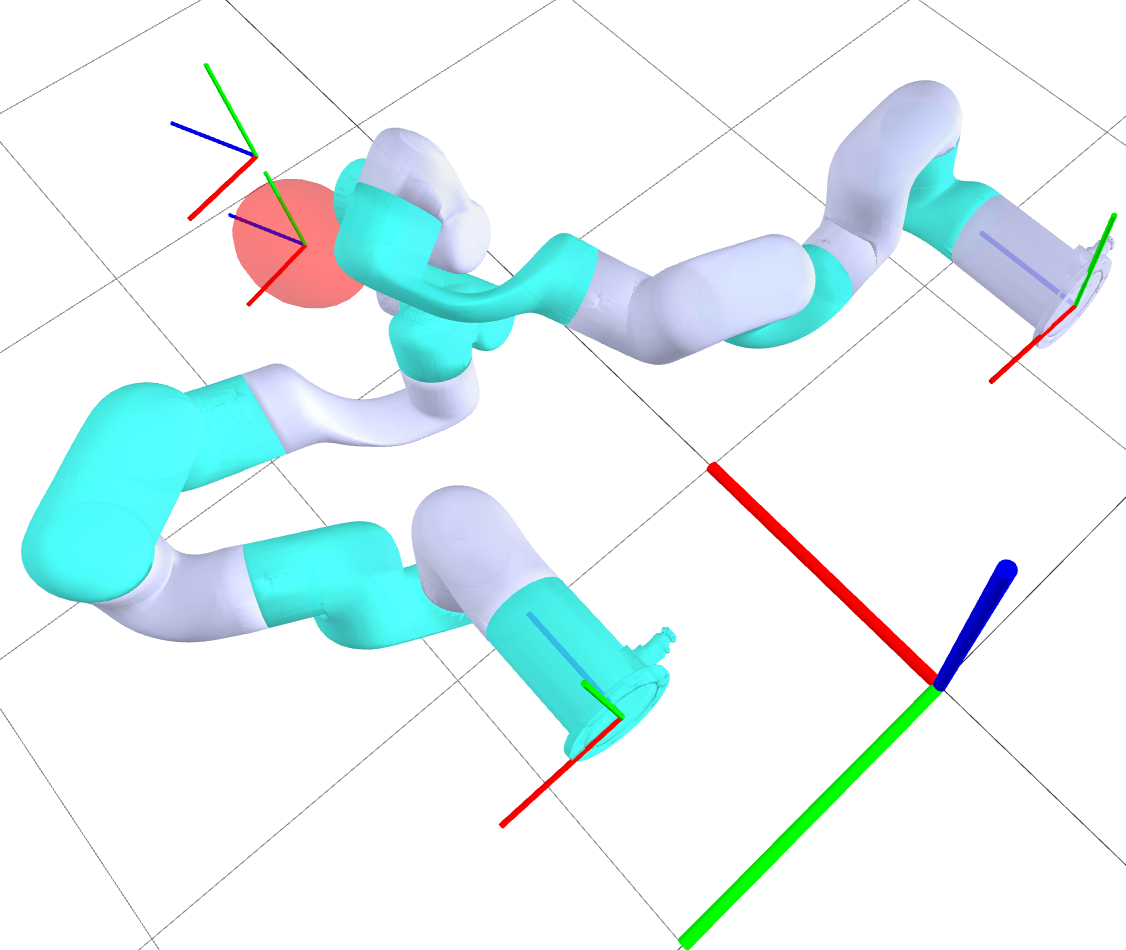}\label{fig:e-design}}
    \subfigure[Baseline (B)]{\includegraphics[width=0.245\textwidth]{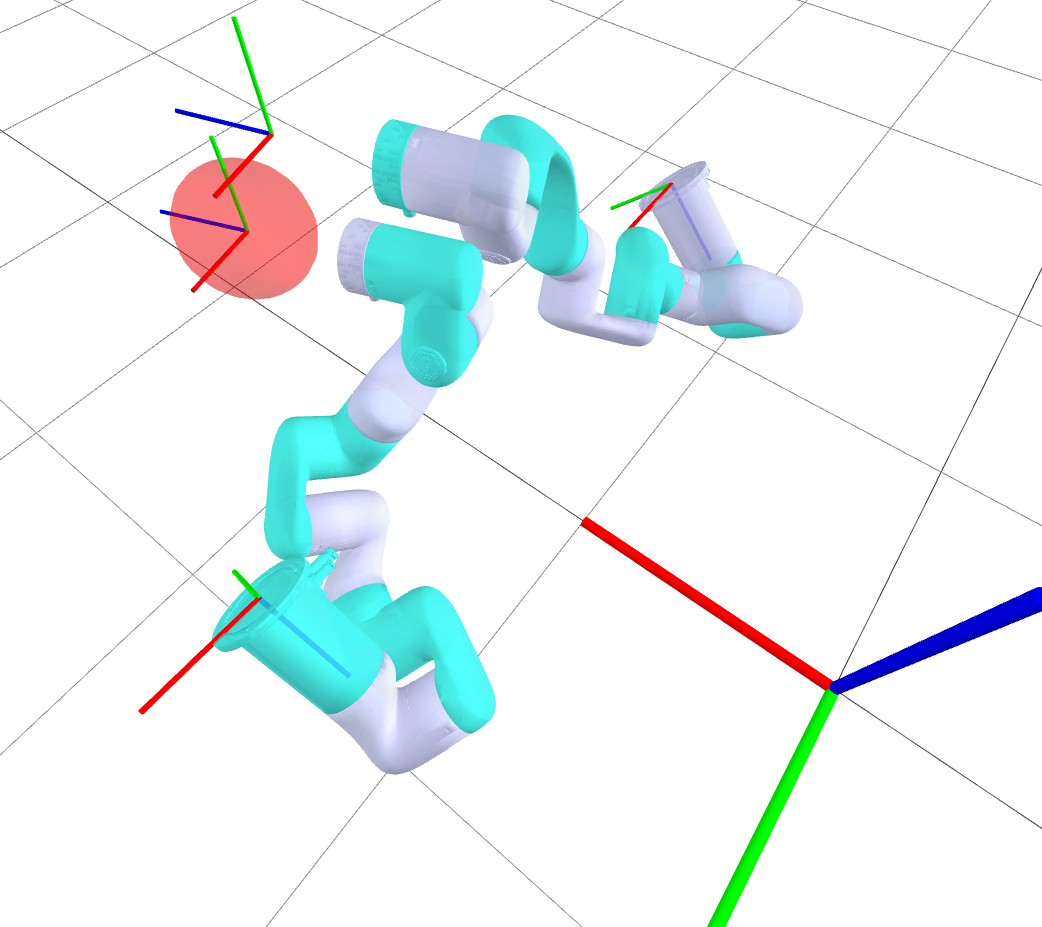}\label{fig:b-design}}
    \caption{The illustrations above showcase the four designs obtained through our optimization method: (a) and (b) from our approach, (c) from the human expert, and (d) from the baseline. Each design is depicted in action, with the right arm, $\mathbf{B}_0(\rho^*)$, reaching for a pepper sample and the left arm, $\mathbf{B}_1(\rho^*)$, cutting the peduncle --- where $\rho^*$ is the realized optimal design (see Table~\ref{tab:final-results}). Solutions (a) and (b) exhibit similar characteristics, featuring a positive camber that enhances the shared manipulator workspace. In contrast, the baseline solution (d) appears to have converged to a local minimum despite the randomized initialization step. A detailed performance comparison is provided in Figure~\ref{fig:design-results-plots}.}\label{fig:design-results-pictures}
\end{figure*}

\section{EXPERIMENTS}\label{sec:iv:experiments}
Our experiments are based on synthetic pepper harvesting scenarios designed to replicate real-world conditions observed in our previous deployments (Figure~\ref{fig:real-conditions-model}). We focus on task-specific design optimization, as formulated in Equation~\ref{eqn:design-task-optimization}, which finds the best solution for a single assigned task. A more general solution would require solving Equation~\ref{eqn:design-expectation-maximization}. We create two design optimization problems: one based on manipulability and another on manipulability density, as defined in Equations~\ref{eqn:yoshikawa} and~\ref{eqn:manip-density}. The four key elements of our design methodology are outlined below.

\subsection{Environment}
Based on the environment model described in Figure~\ref{fig:env-model}, we create a simple generative model to create a synthetic pepper (fruit and peduncle $\SEthree$ frames). We assume a ridge of dimensions $h=0.8$m, $w=0.6$m, and $\ell=0.6$m. The ridge is centered at $(1,0,0)$, extending along the $y$-axis, so that the length spans from $y\in[-\tfrac{1}{2}\ell,\tfrac{1}{2}\ell]$. The ridge is a collision obstacle for the base to the shoulders of the manipulators --- immovable part of the manipulators.

\subsection{Tasks}
The synthetic peppers are generated by first sampling points within the canopy volume, shown in red in Figure~\ref{fig:env-model}. For each sampled point $i$, a corresponding pepper frame $\mathbf{P}_i$ and peduncle frame $\mathbf{C}_i$ are defined as follows:
\begin{align*}
    \mathbf{P}_i &= \mathbf{T}(x_i,y_i,z_i) 
        \cdot \mathbf{R}_x(\tfrac{\pi}{2}+\delta_1)
        \cdot \mathbf{R}_y(\tfrac{\pi}{2}+\delta_2) \\
    \mathbf{C}_i &= \mathbf{P}_i 
        \cdot \mathbf{T}_y(d)
\end{align*}
where $d=0.12$ is the distance from the center of the pepper to the cutting point. The terms $\delta_1$ and $\delta_2$ are Gaussian-distributed variables with zero mean and standard deviations of 0.35 rad and 0.5 rad, respectively, introducing variability in the pepper orientations to simulate real-world conditions.

The task includes 20 peppers and a canopy ridge model, which is an obstacle for the shoulder of the manipulator. While the optimization problem in Equation~\ref{eqn:design-task-optimization} does not produce a generalizing solution, we generate 10 additional sets of 20 targets each to evaluate performance on unseen data. This test offers insight into how well the designs perform beyond the original optimization task.

\subsection{Metrics}
To demonstrate the utility of our design methodology, we define two optimization problems; the first on a manipulability task metrics (Equation~\ref{eqn:yoshikawa}), and the second on manipulability density (Equation~\ref{eqn:manip-density}). Both are computed using self-motion manifolds to ensure all IK solutions are considered.

\subsection{Optimization}
The task metrics and environmental and operational constraints result in a nondifferentiable and highly nonlinear optimization problem. Therefore, we use particle swarm optimization (PSO), a gradient-free global optimization solver with established convergence guarantees~\cite{kennedyParticleSwarmOptimization1995,xuConvergenceAnalysisParticle2018}. We set the solver parameters to an inertial weight of 0.5, acceleration coefficients 2.0, and 10 iterations. To enhance efficiency, we discretized the search space (see Table~\ref{tab:search-spaceparameters}), enabling memoization and ensuring solutions remain within feasible configurations. Under these settings, each optimization run took up to 24 hours on a 20-core i7 processor with 64 GB of RAM.

\subsection{Baseline}
We compared our design optimization results to the expert design shown in Figure~\ref{fig:design-pipeline}, which corresponds to $\rho = (0.175, 0.35, 0.4, 0, 0)$, and an optimization approach based on the manipulability-maximizing controller (MMC)\cite{havilandPurelyReactiveManipulabilityMaximisingMotion2020}. To simplify implementation, we considered only joint limits and canopy collisions. Since the MMC controller operates as a local optimizer, we adopted a random initialization strategy proposed by Balci \etal~\cite{balciOptimalWorkpiecePlacement2023}, which improves robustness by running the solver on $N$ randomly sampled initial guesses instead of a single starting point. We set $N=20$, ensuring the total computation time remained comparable to a single PSO run.

\section{RESULTS}\label{sec:v:results}
Designs are evaluated based on reachability and dexterity—measured by manipulability density (MD) and manipulability (M) (see Figure~\ref{fig:design-results-plots}). For both metrics, the solutions obtained via our proposed methodology, MD and M, achieved the highest performance, with reachability scores of $100\%$ and $95\%$ and dexterity scores of $0.039$ and $0.035$, respectively. In contrast, the baseline solution (B) attained a dexterity score of $0.035$ but reached only $65\%$ of targets, while the expert design (E) reached $70\%$ with the lowest dexterity score of $0.022$. A summary of all the results is presented in Table~\ref{tab:final-results}.

\begin{table}[ht]
    \centering
    \caption{Results for the manipulability density (MD) and manipulability (M) objectives, along with the expert (E) and baseline (B) solutions, are presented. The columns (\textbf{MD}, \textbf{SR}) and (\textbf{MD}${}^*$, \textbf{SR}${}^*$) represent the manipulability density dexterity scores and success rates, evaluated on the optimization task and averaged over the 10 randomly sampled tasks, respectively.
    }\label{tab:final-results}
    \begin{tabular}{|c|c|c|c|c|}
        \hline
& \textbf{Optimal Solution} $\rho^*$ & \textbf{(MD, SR)} & (\textbf{MD}$^*$, \textbf{SR}$^*$) \\ \hline
MD & (0.3, 0.2, 0.5, 90, 25) & (0.039, 100) & (0.0291, 78.2) \\ \hline
M & (0.275, 0.25, 0.65, 90, 25) & (0.035, 95) & (0.0269 , 80.9) \\ \hline
E & (0.175, 0.35, 0.4, 0, 0) & (0.022, 70) & (0.0201, 64.1) \\ \hline
B & (0.65, 0.44, 0.39, 132, -1) & (0.035, 65) & (0.0298 , 60.9) \\ \hline
    \end{tabular}
\end{table}

Although the optimization problem (Equation~\ref{eqn:design-task-optimization}) does not inherently guarantee generalization, performance on ten new tasks --- as represented by the 2$\sigma$-ellipse in Figure~\ref{fig:design-results-plots} --- indicates that our methodology yields designs with improved reachability and dexterity. Despite its intuitive appeal, the expert design trails with reachability and dexterity scores of $64.1\%$ and $0.02$, respectively, and exhibits the highest variance. While the optimization-based solutions show comparable variance, the baseline solution's mean reachability score of $60\%$ suggests it likely became trapped in a local minimum.

Figure~\ref{fig:design-results-pictures} further illustrates that the optimization-based designs are markedly different yet align with what one might expect from a human designer. Although the expert design adheres most closely to a humanoid form, it features the least overlap between the workspace volumes of the two manipulators. Overall, despite the task description being relatively simple, our experiments demonstrate that optimal robot design often eludes intuition-based approaches—underscoring the need for a systematic design methodology to enhance the robot's performance.

\section{CONCLUSION}\label{sec::conclusion}
We introduce a four-part framework that includes a parameterized robot model, a task and environment representation to capture operating conditions, a task metric for performance evaluation, and an optimization algorithm. We demonstrate its effectiveness by applying our methodology to an optimal robot placement problem for a dual-arm pepper-harvesting robot with redundant manipulators. We propose a self-motion manifold-based task metric to comprehensively and accurately assess the best achievable performance. Our results show significant improvements in reach and dexterity compared to human experts and state-of-the-art manipulability maximizing controller techniques. By structuring robot design optimization systematically and in a generalizable manner, our methodology improves design outcomes and provides a framework for analyzing existing robot design processes.

\section*{ACKNOWLEDGMENT}
This work was supported in part by NSF Robust Intelligence 1956163, NSF/USDA-NIFA AIIRA AI Research Institute 2021-67021-35329, and USDA-NIFA LEAP 2024-51181-43291.

\balance
\bibliographystyle{IEEEtran}
\bibliography{structured-design}

\end{document}